%% file: elsarticle-template.tex
\documentclass[preprint, 5p, times, twocolumn]{elsarticle}

\usepackage{amssymb}

\usepackage{comment}	                            
\usepackage{amsmath, multirow, paralist}	        
\usepackage{algorithm}	                            
\usepackage{algorithmic}	                        
\usepackage{graphicx}	                            
\usepackage{caption}                                
\usepackage{subfigure}                              
\usepackage{booktabs}                               
\usepackage{makecell}                               
\usepackage{hyperref}								
\usepackage{setspace}								
\usepackage{url}                                    
\hypersetup{colorlinks=true, urlcolor=black, breaklinks=false}

\journal{Neurocomputing}

\begin{document}

\begin{frontmatter}



\title{Online Parallel Multi-Task Relationship Learning via Alternating Direction Method of Multipliers}

\author[xidian]{Ruiyu Li}
\ead{ruiyli@stu.xidian.edu.cn}
\author[tencent]{Peilin Zhao}
\ead{masonzhao@tencent.com}
\author[xidian]{Guangxia Li}
\ead{gxli@xidian.edu.cn}
\author[uae]{Zhiqiang Xu}
\ead{zhiqiangxu2001@gmail.com}
\author[hit]{Xuewei Li}
\ead{lixuewei@hait.edu.cn}

\affiliation[xidian]{organization={School of Computer Science and Technology, Xidian University},
            country={China}}
   
\affiliation[tencent]{organization={Tencent AI Lab, Tencent Inc},
	city={ShenZhen},
	country={China}}
 
\affiliation[uae]{organization={MBZUAI},
	country={Abu Dhabi, UAE}}

\affiliation[hit]{organization={School of Intelligent Engineering, Henan Institute of Technology},
	country={China}}

\begin{abstract}
Online multi-task learning (OMTL) enhances streaming data processing by leveraging the inherent relations among multiple tasks.
It can be described as an optimization problem in which a single loss function is defined for multiple tasks.
Existing gradient-descent-based methods for this problem might suffer from gradient vanishing and poor conditioning issues.
Furthermore, the centralized setting hinders their application to online parallel optimization, which is vital to big data analytics.
Therefore, this study proposes a novel OMTL framework based on the alternating direction multiplier method (ADMM), a recent breakthrough in optimization suitable for the distributed computing environment because of its decomposable and easy-to-implement nature.
The relations among multiple tasks are modeled dynamically to fit the constant changes in an online scenario.
In a classical distributed computing architecture with a central server, the proposed OMTL algorithm with the ADMM optimizer outperforms SGD-based approaches in terms of accuracy and efficiency.
Because the central server might become a bottleneck when the data scale grows, we further tailor the algorithm to a decentralized setting, so that each node can work by only exchanging information with local neighbors.
Experimental results on a synthetic and several real-world datasets demonstrate the efficiency of our methods.
\end{abstract}


\end{frontmatter}



\input{sec1}
\input{sec2}
\input{sec3}
\input{sec4}
\input{sec5}
\input{sec6}

\bibliographystyle{elsarticle-num}
\bibliography{Ref}

\end{document}

%% file: sec1.tex
\section{Introduction}
\label{sec1}

Online multi-task learning (OMTL) processes related to learning tasks sequentially aim to leverage the correlation among multiple tasks to improve overall performance.
In each online round, the learner receives multiple instances per task, predicts their labels, and then updates the model based on the true labels.
A principal assumption for OMTL is the existence of potential similarities among multiple tasks---the samples of a single task obey a probability distribution similar to the probability distributions of other tasks.
This assumption enables the OMTL to learn several models collaboratively using the shared information among different tasks.
Compared with learning each task separately or treating all tasks as a whole, such a collaborative learning approach can enhance the performance of all tasks together.
OMTL is a real-time, scalable, and continuously adaptive learning method~\cite{DBLP:journals/corr/abs-1210-0473}.
It has been applied in sequential decision making fields that require prompt response, such as online personalized recommendations~\cite{10.1145/3383313.3412236}, targeted display advertising~\cite{10.1145/3447548.3467071}, and sales forecasts for online promotions~\cite{10.1145/3357384.3357823}.

During the past decades, several OMTL algorithms have been proposed, most of which are based on online gradient descent (OGD), such as mirror descent, dual averaging, and their proximal versions~\cite{Li2020, 7959634}.
In particular, OGD is typically used for solving OMTL problems when it is easy to compute the gradient (or sub-gradient) of the online objective, and there are no constraints on the model.
Proximal OGD is usually applied when the regularization term of the model is non-smooth (e.g., $L1$ norm)~\cite{NEURIPS2020_3fe78a8a}.
Its proximal objective frequently enjoys a closed-form solution.
However, for some regularization terms, such as the graph-guided $L1$ norm $\|\mathbf{F}\mathbf{w}\|_1$~\cite{pmlr-v37-zhaob15}, adapting (proximal) OGD methods for distributed online learning settings is non-trivial because sub-gradient methods cannot make $\mathbf{F}\mathbf{w}$ sparse and its proximal objective has no closed-form solution.
Furthermore, the scalability of OGD-based multi-task algorithms deteriorates when the gradient’s dimensionality and the number of tasks increases, making them inadequate for large-scale learning problems.

Unlike OGD methods, the alternating direction multiplier method (ADMM)~\cite{MAL-016} is more applicable to general learning tasks because it does not require the objective to be differentiable.
Specifically, it decomposes the global problem into smaller, easier-to-solve sub-problems suitable for independent workers.
Each worker solves its own sub-problem, which depends only on its own variables.
Subsequently, a server optimizes the global problem by aggregating dual variables from all sub-problems.
Owing to these advantages, ADMM is more suitable for general distributed tasks and is regarded as a viable alternative to OGD for large-scale learning problems~\cite{pmlr-v48-taylor16, 9338293}.

Since ADMM has shown superior ability at optimizing multi-task in the batch learning setting~\cite{10.1145/2939672.2939857, 7551141}, it is attractive to study it in the online scenario, particularly with a distributed computing architecture, so that the learning efficiency can be considerably enhanced by processing multiple tasks in parallel.
Therefore, we propose to perform distributed OMTL using ADMM in this study.
The task-similarity assumption is imposed by decomposing the model-to-learn into two parts: several unique patterns per task and a global pattern shared by all tasks.
The unique patterns are further used to learn the potential relations among tasks on the fly to meet the constant changes in online learning.
We explore the architecture’s two distributed forms, namely, the centralized version with a central server and the relatively decentralized version where all workers involved in solving the optimization problem communicate asynchronously.

The goal of this study is to parallel execute online multi-task learning in distributed computing frameworks, where a task covariance matrix of multiple tasks is exploited to mine the potential relationships among them. 
This is expected to enhance the effectiveness of the proposed method and reduce communication consumption during the optimization process.
We conducted numerical experiments on a synthetic and five real-world datasets~\footnote{Our code is released: \url{https://github.com/Alberta-Lee/NC-24.git}}.
The experimental results demonstrate the effectiveness of this optimization framework.
The rest of this paper is organized as follows:
We outline related works in Section~\ref{sec2} before introducing the OMTL problem setting in Section~\ref{sec3}.
We then deduce the ADMM optimization framework for online parallel multi-classification tasks in Section~\ref{sec4} and analyze its performance experimentally on several datasets in Section~\ref{sec5}.
Finally, we conclude our study in Section~\ref{sec6}.

%% file: sec2.tex
\section{Related Work}
\label{sec2}

\subsection{OMTL}

The field of OMTL has investigated various approaches to address the complexities of simultaneously learning multiple tasks.
Modeling the relations among tasks is crucial for OMTL and directly impacts overall performance.
Existing studies in OMTL typically categorize task relations into two primary types: strong and weak relations.
Strong relations in OMTL often emphasize the similarity of model parameters across tasks.
For example, CMTL~\cite{NIPS2011_a516a87c} assumes that multiple tasks follow a clustered structure, tasks are partitioned into a set of groups based on model parameters, where tasks in the same group are similar to each other.
A new regularizer~\cite{NIPS2006_0afa92fc} based on $(2, 1)$-norm is developed for learning a low-dimensional representation which is shared across a set of multiple related tasks.
To utilize the second-order structure of model parameter, CWMT~\cite{Yang_Zhao_Zhou_Gao_2019} maintains a Gaussian distribution over each model to guide the learning process, where the covariance of the Gaussian distribution is a sum of a local component and a global component that is shared among all the tasks.
Conversely, weak relations consider tasks that may not share strong similarities but still exhibit some degree of relatedness, such as exhibiting similar polarities for the same feature.
For example, ~\cite{10.1007/978-3-031-05936-0_3} explores the convergence properties of optimization methods for multi-convex problems, providing insights to address weakly related tasks using alternating direction methods.
To fully utilize the polarity information of model parameters, SRML~\cite{doi:10.1137/1.9781611977653.ch89} regularizes feature weight signs across tasks to enhance the learning ability of the model.

\subsection{Distributed Optimization}

Distributed optimization plays a crucial role in OMTL, as it makes it possible to process multiple tasks in parallel, thus enhancing the overall performance.
Configuring servers (i.e. centralized vs. decentralized) and making them communicate (i.e. synchronous vs. asynchronous) are fundamental problems for distributed optimization.
It has been theoretically verified that decentralized gradient descent converges to a consistent optimal solution if the expectation of the stochastic delay is bounded and an appropriate step-decreasing strategy is employed.
In addition, their computational complexity is equivalent under certain conditions~\cite{NIPS2017_f7552665, pmlr-v80-lian18a}.
On the other hand, synchronous communication among workers guarantees time-step alignment~\cite{10.1145/3097983.3098136}, whereas the asynchronous approaches have been proven efficient and easy to implement~\cite{7837825}.
However, these optimization methods are based on gradient descent tend to suffer from vanishing gradients, and are sensitive to poor conditioning problems~\cite{10.1145/3292500.3330936} when optimizing a non-convex objective.

As a widely used optimization method, ADMM~\cite{MAL-016} mitigates the gradient vanishing problem by decomposing a complex objective into several simple sub-problems to avoid the chain rule for solving the gradients.
In addition, it is insensitive to inputs and, therefore, immune to poor conditioning~\cite{pmlr-v48-taylor16}.
ADMM has been widely used in multi-task learning~\cite{7552562, 9428542}; however, applying ADMM in OMTL has not yet been thoroughly studied.

%% file: sec3.tex
\section{Problem Setting}
\label{sec3}

In an OMTL problem, we have a set of $K$ parallel tasks whose data $(\mathbf{x}, \mathbf{y})$ all come from the same space $X \times Y$, where $\mathbf{x} \in \mathbb{R}^d, \mathbf{y} \in \mathbb{R}^K$.
For simplicity, we focus on the cases where each is a linear binary classification task, where $X \subset \mathbb{R}^d$, $Y=\{+1, -1\}$, and the model for each task is a vector $\mathbf{w} \in \mathbb{R}^d$, so that its prediction is $\hat{y} = \text{sign} \left(\mathbf{w} \cdot \mathbf{x}\right)$.

Based on these assumptions, an OMTL algorithm works step by step.
Specifically, at the $t$-th round, it receives a group of $K$ instances $\mathbf{x}_t^1, \cdots, \mathbf{x}_t^K$, where $\mathbf{x}_t^k$ is an instance for the $k$-th task.
The algorithm first predicts the labels for each of the tasks as $\hat{y}_t^k = \text{sign} \left( \mathbf{w}_t^k \cdot \mathbf{x}_t^k \right),\  k=1,\cdots, K$. 
It then obtains the true labels $y_t^k$, and suffers a loss $\ell_t^k\left( \mathbf{w}_t^k \right) \triangleq \ell\left( \mathbf{w}_t^k ; (\mathbf{x}_t^k,y_t^k)\right)$, where the loss function $\ell(\cdot)$ is convex, such as hinge loss: $\ell(\mathbf{w};(\mathbf{x}, y)) = \max \left( 0, 1-y(\mathbf{w}^\top\mathbf{x}) \right)$. 
Based on the feedback, the algorithm updates the $K$ classifiers from $\{\mathbf{w}^k_t\}^K_{k=1}$ to $\{\mathbf{w}^k_{t+1}\}^K_{k=1}$ to minimize its loss (plus a regularization term). 
The goal of an OMTL task is to learn a sequence of classifiers $\mathbf{w}_t^1, \cdots, \mathbf{w}_t^K,\ t = 1, \cdots, T$ that achieve the minimum $Regret$ along the entire learning process, where the $Regret$ is defined as:
\begin{equation}\label{eq:regret}
	Regret=\sum_{t=1}^{T} \sum_{k=1}^{K} \ell_t^k(\mathbf{w}_t^k)-\sum_{t=1}^{T} \sum_{k=1}^{K} \ell_t^k(\mathbf{w}_{\ast}^k)
\end{equation}
where $\mathbf{w}_{\ast}^k=\mathop{\mathrm{argmin}}_{\mathbf{w}} \sum_{t=1}^{T} \ell_t^k(\mathbf{w})$ is the optimal classifier for the $k$-th task assuming that we had foresight in all the instances.

The most critical assumption of multi-task learning is that the different tasks are related; thus, the optimal classifiers should be similar in some way.
According to this assumption, we assume that the classifier for each task $\mathbf{w}^k,\ k = 1, \cdots, K$ can be written as: 
\begin{equation}\label{eq:decomposition}
	\mathbf{w}^k = \mathbf{u} + \mathbf{v}^k
\end{equation}
where $\mathbf{u} \in \mathbb{R}^d$ represents the shared pattern of similar tasks, and $\mathbf{v}^k \in \mathbb{R}^d$ catches the unique pattern of a specific task.
Considering some variability across tasks, we simultaneously learn their intrinsic relationships.
We use $\mathbf{\Omega} \in \mathbb{R}^{K \times K}$ to describe the relationship among tasks.

We can thus define the regularized loss function at time $t$ as:
\begin{equation}\label{eq:regularizedLoss}
	\begin{aligned}
		\mathcal{L}_t = &\sum_{k=1}^{K} \ell_{t}^{k} \left( \mathbf{w}_t^k \right) + \frac{\lambda_1}{2} \sum_{k=1}^{K} \Vert \mathbf{v}_t^k \Vert_2^2 + \frac{\lambda_2}{2} \Vert \mathbf{u}_t \Vert_2^2 \\
		&+ \frac{\lambda_3}{2} \text{tr} \left( \mathbf{V}_t\mathbf{V}_t^T \right) + \frac{\lambda_4}{2} \text{tr} \left( \mathbf{V}_t\mathbf{\Omega}_t^{-1}\mathbf{V}_t^T \right)
	\end{aligned}
\end{equation}
where $\mathbf{w}_t^k=\mathbf{u}_t+\mathbf{v}_t^k, \lambda_1, \lambda_2 \textgreater 0, \lambda_3, \lambda_4$ are regularization parameters, and $\mathbf{\Omega}_t \succeq 0$ means that the relationship matrix $\mathbf{\Omega}_t$ is positive semi-definite.
More specifically, $\mathbf{\Omega}_t$ is defined as a task covariance matrix~\cite{10.5555/3023549.3023636}.
The first term in Eq.~\eqref{eq:regularizedLoss} measures the empirical loss on the stream data, the second and third terms penalize the complexity of classifiers from a single task perspective, the fourth term penalizes the complexity of $\mathbf{V}_t$, and the last term measures the relationships among all tasks based on $\mathbf{V}_t$ and $\mathbf{\Omega}_t$.

%% file: sec4.tex
\section{Methodology}
\label{sec4}

We solve the objective for Eq.~\eqref{eq:regularizedLoss} by proposing using the online alternating direction method of the multiplier algorithm~\cite{DBLP:conf/icml/WangB12, DBLP:journals/corr/WangB13a} because it is very scalable to large-scale stream datasets and can be easily distributed to multiple devices.

\begin{algorithm}[htpb]
	\setstretch{1.2}
	\caption{Parallel Multi-task Relationship Learning via ADMM}
	\label{alg:algorithm1}
	\begin{algorithmic}[1]
		
		\STATE \textbf{Initialization: } $\rho > 0, \eta \geq 0, \mathbf{\Omega}_0 = \mathbf{I}_K / K, \mathbf{u}_0 = \mathbf{v}_0^k = \mathbf{w}_0^k = \mathbf{z}_0^k = \mathbf{0} \in \mathbb{R}^d, k = 1, \cdots, K.$
		
		\FOR{$t = 0, \cdots, T$}
		
		\FOR{$k = 1, \cdots, K$}
		
		\STATE Receive a new instance $\mathbf{x}_t^k$; \
		\STATE Make prediction $\hat{y}_t^k=\text{sign} \left( \mathbf{w}_t^k \cdot \mathbf{x}_t^k \right)$; \
		\STATE Receive true label $y_t^k$; \
		\STATE Suffer loss $\ell_{t}^k \left( \mathbf{w}_t^k \right)$; \
		\STATE Update $\mathbf{w}_t^k$ using Eq.~\eqref{eq:solution1-1}; \
		
		\ENDFOR
		
		\STATE Gather $\mathbf{w}_{t+1}^k, \mathbf{z}_t^k$ from all workers; \ 
		\STATE Update $\mathbf{u}_t$ using Eq.~\eqref{eq:solution1-2}; \
		\STATE Send $\mathbf{u}_{t+1}$ to the workers; \
		
		\FOR{$k = 1, \cdots, K$}
		
		\STATE Update $\mathbf{v}_t^k$ using Eq.~\eqref{eq:solution1-3}; \
		\STATE Update $\mathbf{z}_t^k$ using Eq.~\eqref{eq:solution1-4}; \
		
		\ENDFOR
		
		\STATE Update $\mathbf{\Omega}_t$ using Eq.~\eqref{eq:solutionOptimProblem2}; \
		
		\ENDFOR
	\end{algorithmic}
\end{algorithm}

Following the online ADMM setting, we can rewrite our OMTL task at time $t$ as the following optimization problem:
\begin{equation}\label{eq:optimization}
	\begin{aligned}
		\mathop{\min}_{\mathbf{u}_t, \mathbf{W}_t, \mathbf{V}_t, \mathbf{\Omega}_t} \ &\sum_{k=1}^{K} \left( \ell_t^k\left( \mathbf{w}_t^k \right) + \frac{\lambda_1}{2}\Vert \mathbf{v}_t^k \Vert_2^2 \right) + \frac{\lambda_2}{2}\Vert \mathbf{u}_t \Vert_2^2 \\
		&+ \frac{\lambda_3}{2} \text{tr} \left( \mathbf{V}_t\mathbf{V}_t^T \right) + \frac{\lambda_4}{2} \text{tr} \left( \mathbf{V}_t\mathbf{\Omega}_t^{-1}\mathbf{V}_t^T \right) \\
		&+ \eta B_{\phi} \left( \mathbf{W}_{t-1}, \mathbf{W}_t \right) \\
		\mbox{s.t.} \ &\mathbf{w}_t^k - \mathbf{u}_t - \mathbf{v}_t^k = 0, \ k = 1, \cdots, K \\
		&\mathbf{\Omega}_t \succeq 0,\ \text{tr} \left( \mathbf{\Omega}_t \right) = 1
	\end{aligned}
\end{equation}
where $\mathbf{W}_{\ast}=\left[ \mathbf{w}_{\ast}^1, \cdots, \mathbf{w}_{\ast}^K \right], \mathbf{V}_t=\left[ \mathbf{v}_t^1, \cdots, \mathbf{v}_t^K \right] \in \mathbb{R}^{d \times K}$, $\eta \geq 0$ controls the step size.
$B_{\phi}$ is the Bregman divergence defined on a continuously differentiable and strictly convex function $\phi$ to control the distance between $\mathbf{W}_t$ and $\mathbf{W}_{t+1}$.
$B_{\phi} \left( \mathbf{W}_{t-1}, \mathbf{W}_t \right)$ provides a way to quantify and potentially control the variation of the parameter $\mathbf{W}$ from the $t-1$-th round to the $t$-th round.
By choosing the appropriate $\phi$, we can affect the optimization trajectory.
As shown Eq.~\eqref{eq:optimization}, the online distributed multi-task learning problem is a globally consistent optimization.
The first term of Eq.~\eqref{eq:optimization} denotes the objective function partitioned to each worker, $\mathbf{w}_t^k$ and $\mathbf{v}_t^k$ are the local model parameters of worker $k$ at the $t$-th online round and $\mathbf{u}_t$ indicates the global consistency variable.
Each worker independently receives streaming data for parallel training and, through iterative updates, eventually converges to a consistent global model.

At the $t$-th online round, $t = 1, \cdots, T$, we process the optimization problem~\eqref{eq:optimization} in two stages: the first stage deals with the parameters about the learners (or workers), i.e., $\mathbf{w}_t^k, \mathbf{v}_t^k$ and $\mathbf{u}_t$.
When updating these parameters, we follow the ordinary ADMM~\cite{MAL-016} ordering procedure---one can update $\mathbf{w}_t^k$ and $\mathbf{v}_t^k$ for each task in parallel and subsequently update the inter-task shared pattern.
Once we obtain the least parameters (more precisely, $\mathbf{v}_t^k, k = 1, \cdots, K$), the second stage allows us to update the relationship among tasks.
The detailed procedure of the above two stages are as follows:

\subsection{Optimizing $\mathbf{w}_t^k, \mathbf{v}_t^k $ and $\mathbf{u}_t$ When $\mathbf{\Omega}_t$ is Fixed}
Firstly, we fix $\mathbf{\Omega}_t$ and optimize the remaining variables.
This optimization problem is constrained convex, which can be stated as:
\begin{equation}\label{eq:optimProblem1}
	\begin{aligned}
		\mathop{\min}_{\mathbf{u}_t, \mathbf{W}_t, \mathbf{V}_t} \ &\sum_{k=1}^{K} \left( \ell_t^k\left( \mathbf{w}_t^k \right) + \frac{\lambda_1}{2}\Vert \mathbf{v}_t^k \Vert_2^2 \right) + \frac{\lambda_2}{2}\Vert \mathbf{u}_t \Vert_2^2 \\
		&+ \frac{\lambda_3}{2} \text{tr} \left( \mathbf{V}_t\mathbf{V}_t^T \right) + \frac{\lambda_4}{2} \text{tr} \left( \mathbf{V}_t\mathbf{\Omega}_t^{-1}\mathbf{V}_t^T \right) \\
		&+ \eta B_{\phi} \left( \mathbf{W}_{\ast}, \mathbf{W}_t \right) \\
		\mbox{s.t.} \ & \mathbf{w}_t^k - \mathbf{u}_t - \mathbf{v}_t^k = 0, \ k = 1, \cdots, K
	\end{aligned}
\end{equation}

We solve the above problem using ADMM by first deriving the augmented Lagrangian function of problem~\eqref{eq:optimProblem1} as:
\begin{equation}\label{eq:ALMProblem1}
	\begin{aligned}
		L ( \mathbf{W}_t, \mathbf{V}_t, \mathbf{u}_t, \mathbf{z}_t) = &\sum_{k=1}^{K} \left( \frac{\lambda_1}{2}\Vert \mathbf{v}_t^k \Vert_2^2 \right) + \frac{\lambda_2}{2}\Vert \mathbf{u}_t \Vert_2^2 + \frac{\lambda_3}{2} \text{tr} \left( \mathbf{V}_t\mathbf{V}_t^T \right) \\
		&+ \sum_{k=1}^{K} \left( \ell_{t}^{k} \left( \mathbf{w}_t^k \right) + \mathbf{z}_t^k \cdot \left( \mathbf{w}_t^k-\mathbf{u}_t-\mathbf{v}_t^k \right) \right) \\
		&+ \sum_{k=1}^{K} \left( \frac{\rho}{2}\Vert \mathbf{w}_t^k-\mathbf{u}_t-\mathbf{v}_t^k \Vert_2^2  \right) \\
		&+ \frac{\lambda_4}{2} \text{tr} \left( \mathbf{V}_t\mathbf{\Omega}_t^{-1}\mathbf{V}_t^T \right) + \eta B_{\phi} \left( \mathbf{W}_{\ast}, \mathbf{W}_t \right)
	\end{aligned}
\end{equation}
where $\mathbf{z}_t^k \in \mathbb{R}^d$ are the dual variables and $\rho > 0$ is the penalty parameter.

Subsequently, according to the online ADMM algorithm, our algorithm comprises updates of the primal variables $\mathbf{W}_t,\ \mathbf{V}_t,\ \mathbf{u}_t$ and dual variables $\mathbf{z}_t$.

\textbf{Updating $\mathbf{W}_t$.}
The update of $\mathbf{W}_t$ can be written as:
\begin{equation}\label{eq:optimProblem1-1}
	\begin{aligned}
		\mathbf{W}_{t+1} = \mathop{\mathrm{argmin}}_{\mathbf{W}_t} \ &L \left( \mathbf{W}_t, \mathbf{V}_t, \mathbf{u}_t, \mathbf{Z}_t \right) \\
		= \mathop{\mathrm{argmin}}_{\mathbf{W}_t} \ &\sum_{k=1}^{K} \left( \ell_{t}^{k} \left( \mathbf{w}_t^k \right) + \mathbf{z}_t^k \cdot \left( \mathbf{w}_t^k-\mathbf{u}_t-\mathbf{v}_t^k \right) \right) \\
		+ &\sum_{k=1}^{K} \left( \frac{\rho}{2}\Vert  \mathbf{w}_t^k-\mathbf{u}_t-\mathbf{v}_t^k \Vert_2^2 \right) + \eta B_{\phi} \left( \mathbf{W}_{\ast}, \mathbf{W}_t \right)
	\end{aligned}
\end{equation}

However, it is challenging to solve the closed-form solution of the above optimization problem~\eqref{eq:optimProblem1-1} for the hinge loss function.
Thus, we adopt the first-order approximation of hinge loss:
\begin{equation}\label{eq:firstOrderApprox}
	\begin{aligned}
		\ell_t^k \left( \mathbf{w} \right) \approx \ell_t^k \left( \mathbf{w}_t^k \right) + \nabla \ell_t^k \left( \mathbf{w}_t^k \right)^T \left( \mathbf{w} - \mathbf{w}_t^k \right)
	\end{aligned}
\end{equation}

Furthermore, we consider $ B_{\phi} \left( \mathbf{w}, \mathbf{u} \right) = \frac{1}{2} \Vert \mathbf{w} - \mathbf{u} \Vert_2^2$ for simplicity so that
\begin{equation}\label{eq:bregmanDivergence}
	\begin{aligned}
		B_{\phi} \left( \left[ \mathbf{w}_{\ast}^1, \cdots, \mathbf{w}_{\ast}^K \right], \left[ \mathbf{w}_t^1, \cdots, \mathbf{w}_t^K \right] \right) = \frac{1}{2} \sum_{k=1}^{K} \Vert \mathbf{w}_{\ast}^k - \mathbf{w}_t^k \Vert_2^2
	\end{aligned}
\end{equation}

Combining the above equations gives an approximate solution of problem~\eqref{eq:optimProblem1-1} as:
\begin{equation}\label{eq:solution1-1}
	\begin{aligned}
		\mathbf{w}_{t+1}^k = &\frac{\eta}{\rho+\eta} \mathbf{w}_t^k + \frac{\rho}{\rho+\eta} \left( \mathbf{u}_t + \mathbf{v}_t^k \right) \\
		- &\frac{1}{\rho+\eta} \left( \nabla \ell_t^k \left( \mathbf{w}_t^k \right) + \mathbf{z}_t^k \right)
	\end{aligned}
\end{equation}

\begin{algorithm}
	\setstretch{1.2}
	\caption{Decentralized Framework}
	\label{alg:algorithm2}
	\begin{algorithmic}[1]
		
		\STATE \textbf{Initialization:} $\rho > 0, \eta \geq 0, \mathbf{\Omega}_0 = \mathbf{I}_K / K, \mathbf{u}_0 = \mathbf{v}_0^k = \mathbf{w}_0^k = \mathbf{z}_0^k = \mathbf{0} \in \mathbb{R}^d, k = 1, \cdots, K.$
		
		\FOR{$t = 0, \cdots, T$}
		
		\FOR{$k = 1, \cdots, K$}
		
		\STATE Receive a new instance $\mathbf{x}_t^k$; \
		\STATE Make prediction $\hat{y}_t^k=\text{sign} \left( \mathbf{w}_t^k \cdot \mathbf{x}_t^k \right)$; \
		\STATE Receive true label $y_t^k$; \
		\STATE Suffer loss $\ell_{t}^k \left( \mathbf{w}_t^k \right)$;
		\STATE Update $\mathbf{w}_t^k$;
		
		\ENDFOR
		
		\STATE Gather $\mathbf{w}_{t+1}^k, \mathbf{z}_t^k$ from its 1-hop neighbors; \
		\STATE Update $\mathbf{u}_t$; \
		\STATE Send $\mathbf{u}_{t+1}$ to its neighbors; \
		
		\FOR{$k = 1, \cdots, K$}
		
		\STATE Update $\mathbf{v}_t^k$;
		\STATE Update $\mathbf{z}_t^k$;
		
		\ENDFOR
		
		\STATE Update $\mathbf{\Omega}_t$;
		
		\ENDFOR
		
	\end{algorithmic}
\end{algorithm}

\begin{table*}
	\caption{Statistics of datasets used in the experiment.}
	\label{tab:tab1}
	\centering
	\begin{tabular}{lcccccc}
		\toprule
		& \text{Synthetic}   & \text{Tweet Eval}   & \text{Multi-Lingual}   & \text{Chem}   & \text{Landmine}   & \text{MNIST} \\
		\midrule
		\text{Num of Task}              & \text{5}           & \text{3}            & \text{5}               & \text{6}      & \text{29}     & \text{5}       \\
		\text{Num of Feature Dimension} & \text{9}           & \text{512}          & \text{512}             & \text{64}     & \text{9}    & \text{512}     \\
		\text{Total Sample Count}       & \text{50000}       & \text{31671}        & \text{187092}          & \text{7926}   & \text{14820}  & \text{60000}   \\
		\text{Max Sample Count}         & \text{10000}       & \text{14100}        & \text{84000}           & \text{4110}   & \text{690}   & \text{12660}   \\
		\text{Min Sample Count}         & \text{10000}       & \text{4601}         & \text{2022}            & \text{188}    & \text{445}   & \text{11344}   \\
		\text{Positive Ratio}           & \text{0.50}        & \text{0.41}         & \text{0.54}            & \text{0.50}   & \text{0.06}  & \text{0.51}    \\
		\bottomrule
	\end{tabular}
\end{table*}

\textbf{Updating $\mathbf{V}_t$.}
The unique pattern $\mathbf{V}_t$ for each task can be updated as:
\begin{equation}\label{eq:optimProblem1-0}
	\begin{aligned}
		\mathbf{V}_{t+1} = \mathop{\mathrm{argmin}}_{\mathbf{V}_t} \ &L \left( \mathbf{W}_{t+1}, \mathbf{V}_t, \mathbf{u}_t, \mathbf{Z}_t \right) \\
		= \mathop{\mathrm{argmin}}_{\mathbf{V}_t} &\sum_{k=1}^{K} \left( \frac{\lambda_1}{2}\Vert \mathbf{v}_t^k \Vert_2^2 + \mathbf{z}_t^k \cdot \left( \mathbf{w}_{t+1}^k-\mathbf{u}_t-\mathbf{v}_t^k \right) \right) \\
		+ &\sum_{k=1}^{K} \left( \frac{\rho}{2}\Vert  \mathbf{w}_{t+1}^k-\mathbf{u}_t-\mathbf{v}_t^k \Vert_2^2 \right) \\
		+ &\frac{\lambda_3}{2} \text{tr} \left( \mathbf{V}_t\mathbf{V}_t^T \right) + \frac{\lambda_4}{2} \text{tr} \left( \mathbf{V}_t\mathbf{\Omega}_t^{-1}\mathbf{V}_t^T \right)
	\end{aligned}
\end{equation}

With the careful deduction of Eq.~\eqref{eq:optimProblem1-0}, we can derive the following solution:
\begin{equation}\label{eq:solution1-3}
	\begin{aligned}
		\mathbf{v}_{t+1}^k = &\frac{\lambda_2 \left( \mathbf{z}_t^k + \rho \mathbf{w}_{t+1}^k \right) }{\lambda_2 \left(\lambda_1 + \lambda_3 + \rho \right) + \rho K \left(\lambda_1 + \lambda_3 \right) } \\
		&+ \frac{\lambda_4}{2} \left[ \mathbf{V}_t\mathbf{\Omega}_t^{-1} + \mathbf{V}_t \left( \mathbf{\Omega}_t^{-1} \right)^T \right]_{:, k}
	\end{aligned}
\end{equation}
where $\left[ \mathbf{V}_t\mathbf{\Omega}_t^{-1} + \mathbf{V}_t \left( \mathbf{\Omega}_t^{-1} \right)^T \right]_{:, k}$ denotes the $k$-th column of the matrix.

\textbf{Updating $\mathbf{u}_t$.}
Simultaneously, the shared pattern $\mathbf{u}_t$ of the similar tasks can be updated as:
\begin{equation}\label{eq:optimProblem1-3}
	\begin{aligned}
		\mathbf{u}_{t+1} = \mathop{\mathrm{argmin}}_{\mathbf{u}_t} \ &L \left( \mathbf{W}_{t+1}, \mathbf{V}_t, \mathbf{u}_t, \mathbf{Z}_t \right) \\
		=\mathop{\mathrm{argmin}}_{\mathbf{u}_t} &\sum_{k=1}^{K} \left( \mathbf{z}_t^k \cdot \left( \mathbf{w}_{t+1}^k-\mathbf{u}_t-\mathbf{v}_t^k \right) \right) \\
		+&\sum_{k=1}^{K} \left( \frac{\rho}{2}\Vert  \mathbf{w}_{t+1}^k-\mathbf{u}_t-\mathbf{v}_t^k \Vert_2^2 \right) + \frac{\lambda_2}{2}\Vert \mathbf{u}_t \Vert_2^2
	\end{aligned}
\end{equation}
It is easy to derive the solution for $\mathbf{u}_t$ as:
\begin{equation}\label{eq:solution1-2}
	\begin{aligned}
		\mathbf{u}_{t+1} = \frac{\left(\lambda_1 + \lambda_3\right) \sum_{k=1}^{K} \left( \mathbf{z}_t^k + \rho \mathbf{w}_{t+1}^k \right) }{\left(\lambda_1 + \lambda_3\right)\left(\lambda_2 + \rho K\right) + \lambda_2 \rho} 
	\end{aligned}
\end{equation}

\textbf{Updating $\mathbf{z}_t$.}
Finally, the dual variables are updated as:
\begin{equation}\label{eq:solution1-4}
	\begin{aligned}
		\mathbf{z}_{t+1}^k = \mathbf{z}_{t}^k + \rho \left( \mathbf{w}_{t+1}^k - \mathbf{u}_{t+1} - \mathbf{v}_{t+1}^k \right)
	\end{aligned}
\end{equation}

Note that the update of $\mathbf{w}_t^k$ and $\mathbf{v}_t^k$ can be paralleled for each task.
Thus, it is easy to solve the optimization problem in a centralized network with one central server node, and $K$ workers connect to the server.

\subsection{Optimizing $\mathbf{\Omega}_t$ When $\mathbf{V}_t$ is Fixed}
Finally, we optimize the variable $\mathbf{\Omega}_t$ while fixing all the other variables. 
This optimization problem can be expressed as the following constrained one: 
\begin{equation}\label{eq:optimProblem2}
	\begin{aligned}
		\mathop{\min}_{\mathbf{\Omega}_t} \ &\text{tr} \left( \mathbf{\Omega}_t^{-1}\mathbf{V}_t^T\mathbf{V}_t \right) \\
		\mbox{s.t.} \ &\mathbf{\Omega}_t \succeq 0, \ \text{tr} \left( \mathbf{\Omega}_t \right) = 1
	\end{aligned}
\end{equation}
Subsequently, denote $\mathbf{A}_t = \mathbf{V}_t^T\mathbf{V}_t$, and we can derive the following inequalities:
\begin{equation}\label{eq:analyoptimProblem2}
	\begin{aligned}
		\text{tr} \left( \mathbf{\Omega}_t^{-1}\mathbf{A}_t \right) =\ &\text{tr} \left( \mathbf{\Omega}_t^{-1}\mathbf{A}_t \right) \text{tr}\left( \mathbf{\Omega}_t \right) \\
		\geq \ &\left( \text{tr} \left( \mathbf{A}_t^{\frac{1}{2}} \right) \right)^2
	\end{aligned}
\end{equation}
where the first equality holds because of the last constraint in problem~\eqref{eq:analyoptimProblem2}, and the last inequality holds because of the Cauchy-Schwarz inequality for the Frobenius norm.
Moreover, $\text{tr} \left( \mathbf{\Omega}_t^{-1}\mathbf{A}_t \right)$ attains its minimum value $\left( \text{tr} \left( \mathbf{A}_t^{1/2} \right) \right)^2$ if, and only if, $\mathbf{\Omega}_t^{-1/2}\mathbf{A}_t^{1/2} = a \mathbf{\Omega}_t^{1/2}$ for some constant $a$, $ \text{tr} \left( \mathbf{\Omega}_t \right) = 1$.
Therefore, we can obtain the analytical solution for optimization problem~\eqref{eq:optimProblem2}:
\begin{equation}\label{eq:solutionOptimProblem2}
	\begin{aligned}
		\mathbf{\Omega}_t = \frac{\left(\mathbf{V}_t^T \mathbf{V}_t\right)^{1/2}}{\text{tr} \left( \left(\mathbf{V}_t^T \mathbf{V}_t \right)^{1/2} \right)}
	\end{aligned}
\end{equation}

Furthermore, we set the initial value of $\mathbf{\Omega}_0$ to $\mathbf{I}_K / K$, corresponding to the assumption that all tasks are initially unrelated.
After learning the optimal values of $\mathbf{w}_t^k, \mathbf{v}_t^k, \mathbf{u}_t$ and $\mathbf{\Omega}_t$, we can predict the following set of instances, $\{\mathbf{x}_t^k\}^K_{k=1}$.

Finally, our framework for centralized distributed OMTL can be summarized as in Algorithm~\ref{alg:algorithm1}.
Note that lines \textit{10}-\textit{12} and \textit{17} are performed by the central server.
Similarly, we extend our framework to the decentralized OMTL setting, summarized in Algorithm~\ref{alg:algorithm2}.
According to Eq.~\eqref{eq:solutionOptimProblem2}, updating the task relationship matrix $\mathbf{\Omega}_t$ requires the latest $\mathbf{v}_t^k$ on all workers, which is easy to implement in our centralized architecture with the central server.
We can only obtain the latest $\mathbf{v}_t^k$ from 1-hop neighbor workers in the decentralized framework.
These workers can obtain the remaining $\mathbf{v}_t^k$ by accessing their neighbors, thus aggregating all the $\mathbf{v}_t^k$ of all the workers in the network.
Thus we can still update $\mathbf{\Omega}_t$ by Eq.~\eqref{eq:solutionOptimProblem2}.

%% file: sec5.tex
\section{Experimental Results}
\label{sec5}

\subsection{Experimental Testbeds}

\begin{figure*}[htb]
	\centering
	\subfigure[Synthetic]{
		\includegraphics[width=2.1in]{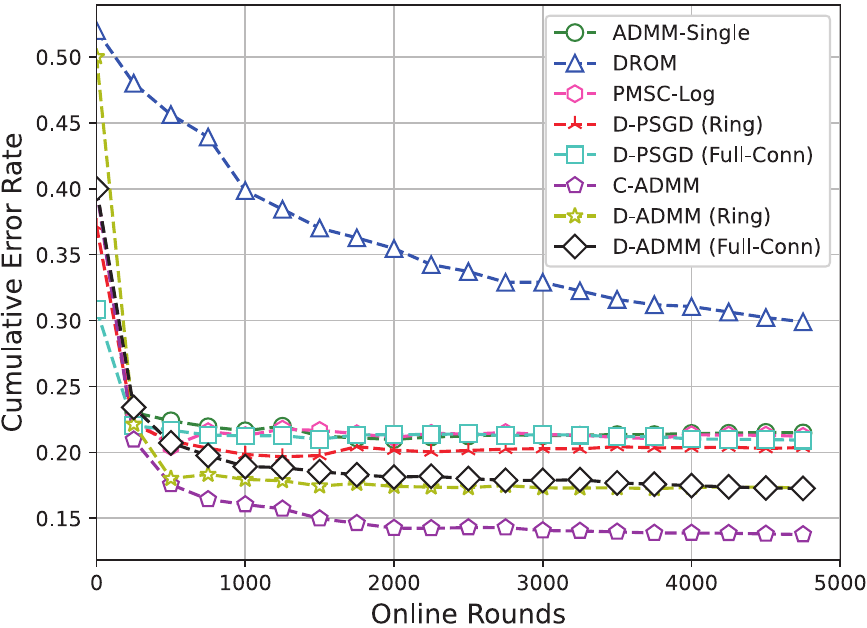}
		\label{fig1a}
	}
	\quad
	\subfigure[Tweet Eval]{
		\includegraphics[width=2.1in]{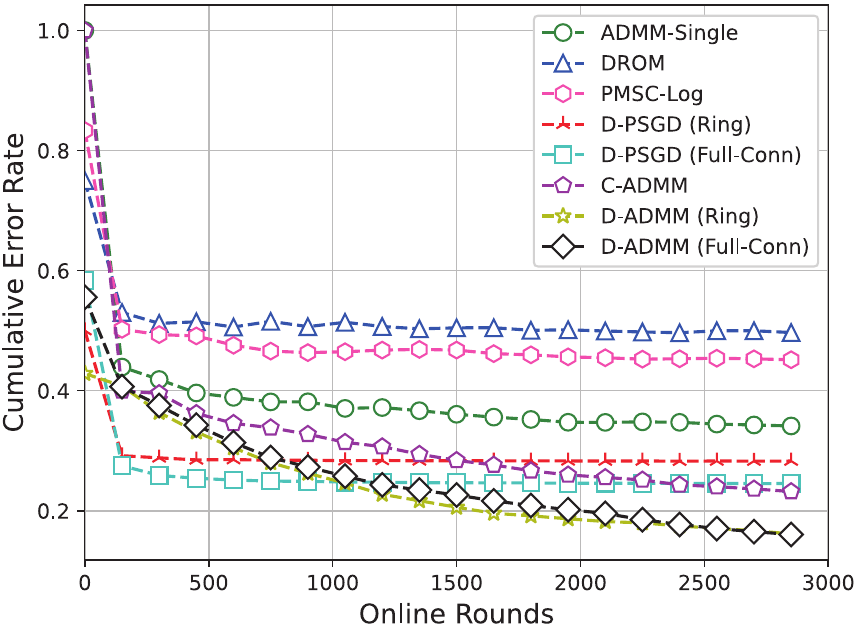}
		\label{fig1b}
	}
	\quad
	\subfigure[Multi-Lingual]{
		\includegraphics[width=2.1in]{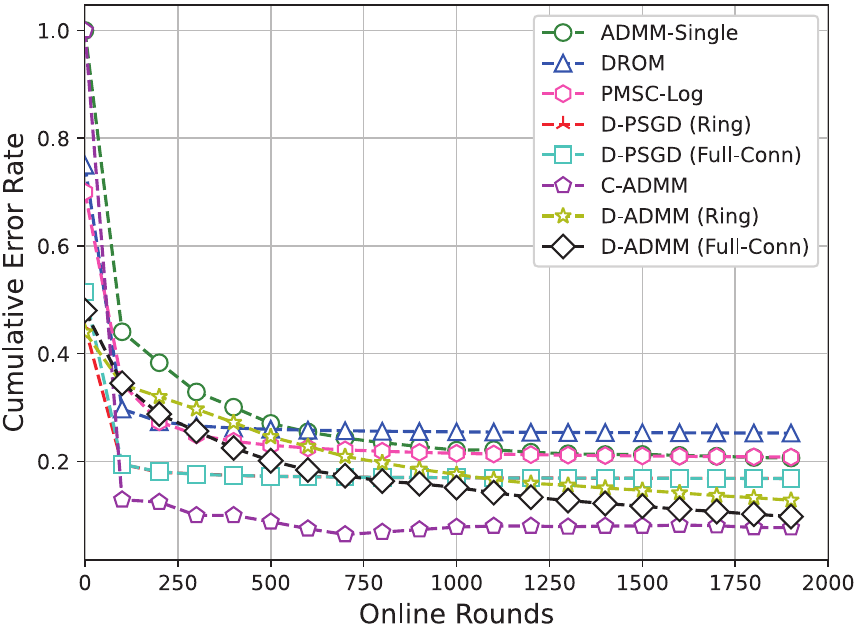}
		\label{fig1c}
	}
	
	\subfigure[Chem]{
		\includegraphics[width=2.1in]{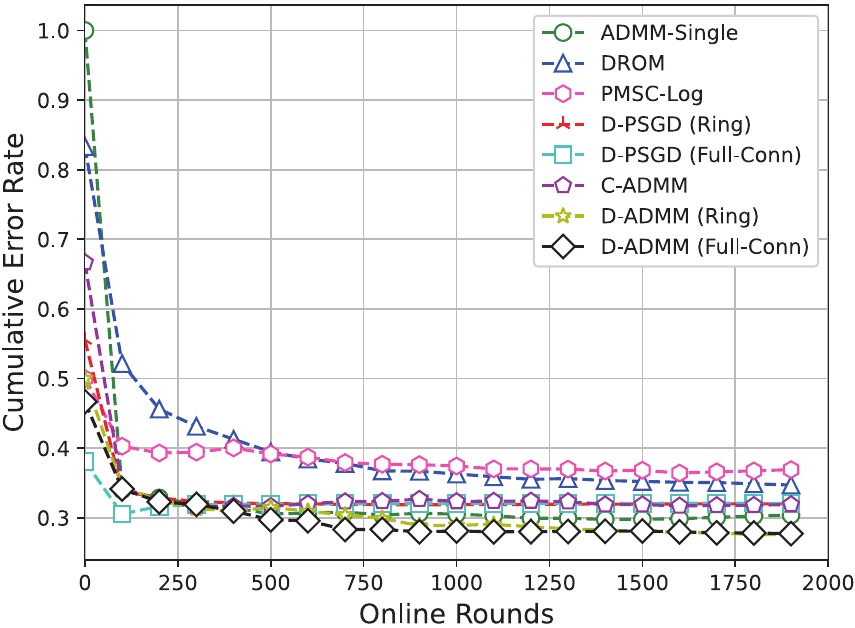}
		\label{fig1d}
	}
	\quad
	\subfigure[Landmine]{
		\includegraphics[width=2.1in]{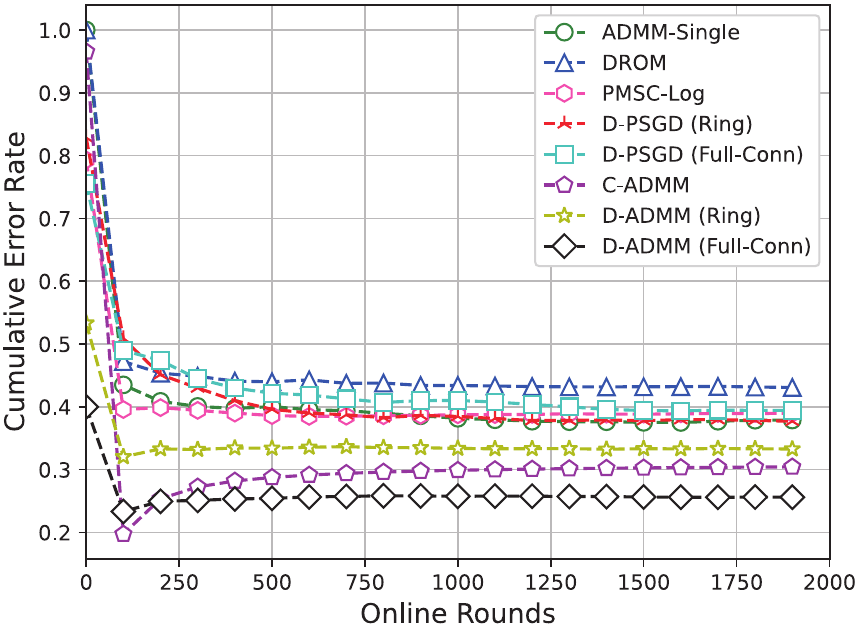}
		\label{fig1e}
	}
	\quad
	\subfigure[MNIST]{
		\includegraphics[width=2.1in]{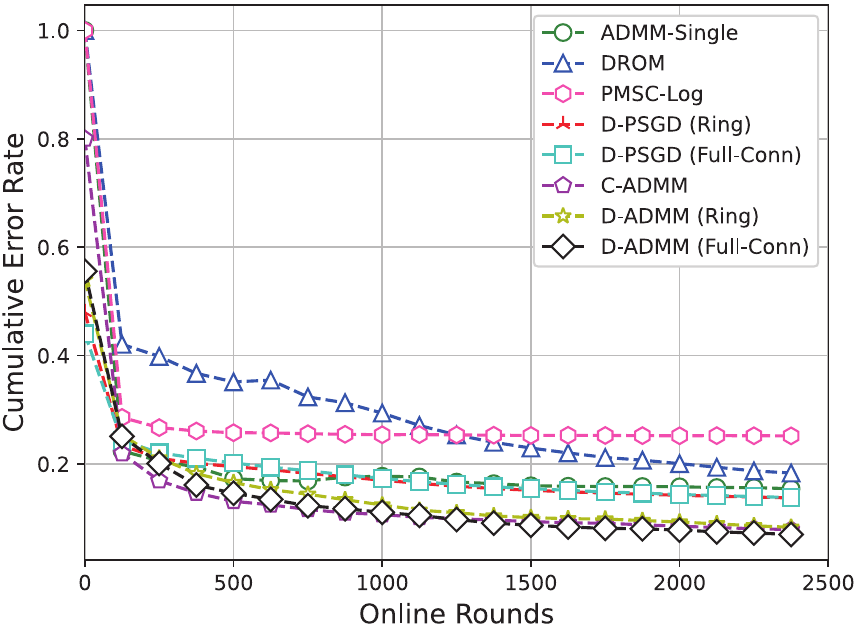}
		\label{fig1f}
	}
	\caption{Averaged variations of the cumulative error rate over all tasks along the entire online learning process on six datasets.}
	\label{fig1}
\end{figure*}

\begin{table*}
	\caption{Experimental results of the averaged error rate of all tasks when algorithms reach the last learning round.}
	\label{tab:tab2}
	\centering
	\begin{tabular}{lcccccc}
		\toprule
		& \text{Synthetic} & \text{Tweet Eval}    & \text{Multi-Lingual} & \text{Chem}   &   \text{Landmine}     &  \text{MNIST}   \\
		\midrule
		\text{ADMM-Single}       & $0.214$          & $0.341$              & $0.206$           & $0.304$          & $0.379$          & $0.154$         \\
		\text{DROM}              & $0.299$          & $0.497$              & $0.253$           & $0.347$          & $0.431$          & $0.183$         \\
		\text{PMSC-Log}          & $0.212$          & $0.452$              & $0.208$           & $0.369$          & $0.389$          & $0.252$         \\
		\text{D-PSGD (Ring)}      & $0.204$          & $0.283$              & $0.168$           & $0.320$          & $0.377$          & $0.137$         \\
		\text{D-PSGD (Full-Conn)} & $0.209$          & $0.245$              & $0.168$           & $0.320$          & $0.394$          & $0.124$         \\
		\text{C-ADMM}            & $\mathbf{0.138}$ & $0.232$              & $\mathbf{0.077}$  & $0.319$         & $0.304$          & $0.079$         \\
		\text{D-ADMM (Ring)}      & $0.173$          & $0.161$              & $0.128$           & $\mathbf{0.276}$ & $0.332$ & $0.070$         \\
		\text{D-ADMM (Full-Conn)} & $0.173$          & $\mathbf{0.160}$     & $0.098$           & $0.277$          & $\mathbf{0.256}$          & $\mathbf{0.051}$\\
		\bottomrule
	\end{tabular}
\end{table*}

We use a synthetic and five real-world datasets to evaluate our methods.
The real-world datasets are sourced from three typical multi-task learning applications: sentiment analysis, small molecule classification, and image classification.

\begin{itemize}
	\item \textbf{Synthetic Dataset}~\cite{6576111}.
	It contains five binary classification tasks whose similarities are controlled by a set of parameters.
	The basic problem is discriminating two classes in a two-dimensional plane with a non-linear decision boundary.
	Changing the parameter rotates the decision boundary to create tasks that look similar but have subtle differences.
	
	\item \textbf{Tweet Eval Dataset}\footnote{https://huggingface.co/datasets/tweet\_eval}.
	It contains three Tweet sentiment classification tasks, i.e. \textit{hate}, \textit{irony}, and \textit{offensive}---all are negative emotions; thus, it makes sense to believe there are commonalities in Tweet texts.
	
	\item \textbf{Multi-Lingual Dataset}\footnote{https://huggingface.co/datasets/tyqiangz/multilingual-sentiments}.
	It collects product reviews from Amazon in five languages: \textit{Chinese}, \textit{English}, \textit{Japanese}, \textit{Indonesian}, and \textit{Malay}.
	Each language contains positive and negative reviews.
	
	\item \textbf{Chem Dataset}\footnote{https://chrsmrrs.github.io/datasets/docs/datasets/\label{chrsmrrs}}.
	It contains six small molecule active classification tasks, such as distinguishing between classes of HIV molecules (active vs. inactive).
	
    \item \textbf{Landmine Dataset}.
    It includes twenty-nine landmine fields. 
    For each field, every sample in the dataset consists of nine features and a binary label indicating whether the corresponding location contains landmines.
	
	\item \textbf{MNIST Dataset}.
	It comprises handwritten digits for image recognition.
	Following the setup in~\cite{pmlr-v51-wang16d}, we create five binary classification tasks as \textit{0} vs. \textit{5}, \textit{1} vs. \textit{6}, \textit{2} vs. \textit{7}, \textit{3} vs. \textit{8}, and \textit{4} vs. \textit{9}.
	Each image is represented by a 512-dimensional vector after processing by using a pre-trained ResNet18 model.
\end{itemize}

We use a transformer for the two sentiment analysis datasets to convert the raw text into vectors of dimension $512$.
The graph embedding~\cite{Morris_Ritzert_Fey_Hamilton_Lenssen_Rattan_Grohe_2019} is applied to generate the corresponding embedding vectors for each molecule for the two small molecule datasets. 
Table~\ref{tab:tab1} summarizes their task numbers, sample sizes, feature counts and class distribution.

\begin{table*}
	\caption{The number of learning rounds (left-side of a column) and the averaged time consumption per round (in milliseconds, right-side of a column) for each algorithm to reach the specified accuracy; the empty cell in the table indicates that the algorithm fails to achieve the specified accuracy anyhow.}
	\label{tab:tab3}
	\centering
	\begin{tabular}{l|cc|cc|cc|cc|cc|cc}
		\toprule
		\makecell{\\ \text{Target accuracy}} & \multicolumn{2}{c}{\makecell{\text{Synthetic} \\ $0.75$}} & \multicolumn{2}{|c}{\makecell{\text{Tweet Eval} \\ $0.60$}} & \multicolumn{2}{|c}{\makecell{\text{Multi-Lingual} \\ $0.70$}} & \multicolumn{2}{|c}{\makecell{\text{Chem} \\ $0.60$}} & \multicolumn{2}{|c}{\makecell{\text{Landmine} \\ $0.55$}} & \multicolumn{2}{|c}{\makecell{\text{MNIST} \\ $0.70$}} \\
		\midrule
		\text{ADMM-Single}        & $267$ & $0.24$     & $485$ & $0.04$       & $507$ & $0.04$        & $121$ & $0.03$      & $139$ & $0.03$      & $204$  & $0.04$     \\
		\text{DROM}               &       & $4.29$     &       & $0.19$       & $139$ & $0.44$        & $502$ & $0.44$      & $247$ & $0.23$      & $1057$ & $0.67$     \\
		\text{PMSC-Log}           & $188$ & $0.26$     &       & $1.90$       & $232$ & $3.03$        & $187$ & $0.76$      & $132$ & $0.18$      & $220$  & $3.21$     \\
		\text{D-PSGD (Ring)}      & $206$ & $3.28$     & $155$ & $267.37$     & $105$ & $1259.42$     & $109$ & $26.34$     & $233$ & $35.87$     & $185$  & $184.27$   \\
		\text{D-PSGD (Full-Conn)} & $204$ & $5.79$     & $157$ & $293.05$     & $110$ & $1479.16$     & $37$  & $48.90$     & $276$& $65.01$     & $184$  & $233.85$   \\
		\text{C-ADMM}             & $261$ & $2.01$     & $190$ & $1.93$       & $114$ & $2.44$        & $116$ & $2.56$      & $97$ & $6.26$      & $196$  & $2.61$     \\
		\text{D-ADMM (Ring)}      & $277$ & $1.03$     & $189$ & $17.63$      & $344$ & $35.83$       & $94$  & $15.05$     & $78$ & $20.99$      & $188$  & $17.76$    \\
		\text{D-ADMM (Full-Conn)} & $275$ & $1.05$     & $190$ & $32.60$      & $249$ & $59.57$       & $86$  & $17.57$     & $32$ & $47.01$      & $189$  & $30.49$    \\
		\bottomrule
	\end{tabular}
\end{table*}

\begin{table*}
	\caption{Ablation study results showing the effect of the proposed relationship learning for OMTL problems.}
	\label{tab:tab4}
	\centering
	\begin{tabular}{l|c|cc|cc|cc}
		\toprule
		\multirow{2}{*}{}    & \multicolumn{1}{c|}{Indpt} & \multicolumn{2}{c|}{C-ADMM}     & \multicolumn{2}{c|}{D-ADMM (Full-Conn)} & \multicolumn{2}{c}{D-ADMM (Ring)} \\
		&                            & \text{W/O RL} & \text{With RL}  & \text{W/O RL} & \text{With RL}          & \text{W/O RL} & \text{With RL}\\
		\midrule
		\text{Synthetic}     & $0.215$                    & $0.162$      & $\mathbf{0.138}$ & $0.183$       & $\mathbf{0.173}$        & $0.191$      & $\mathbf{0.173}$ \\
		\text{Tweet Eval}    & $0.341$                    & $0.356$      & $\mathbf{0.232}$ & $0.320$       & $\mathbf{0.160}$        & $0.324$      & $\mathbf{0.161}$ \\
		\text{Multi-Lingual} & $0.206$                    & $0.099$      & $\mathbf{0.077}$ & $0.114$       & $\mathbf{0.097}$        & $0.165$      & $\mathbf{0.128}$ \\
		\text{Chem}          & $0.352$                    & $0.344$      & $\mathbf{0.319}$ & $0.373$       & $\mathbf{0.277}$        & $0.343$      & $\mathbf{0.276}$ \\
		\text{Landmine}           & $0.379$                    & $0.348$      & $\mathbf{0.304}$ & $0.391$       & $\mathbf{0.256}$        & $0.393$      & $\mathbf{0.332}$ \\
		\text{MNIST}         & $0.154$                    & $0.106$      & $\mathbf{0.079}$ & $0.150$       & $\mathbf{0.051}$        & $0.097$      & $\mathbf{0.070}$ \\
		\bottomrule
	\end{tabular}
\end{table*}

\subsection{Benchmark Setup.}

We refer to the proposed distributed OMTL with ADMM with a central server as \textit{C-ADMM} and its decentralized variant as \textit{D-ADMM}.
Two topologies abbreviated as \textit{Ring} and \textit{Full-Conn} are considered, where the former represents a ring network, and the latter connects each worker to others in the network. 
They are benchmarked against four classical OMTL methods as follows.

\begin{itemize}
	\item \textbf{ADMM-Single}.
	It employs the ADMM algorithm to train a single model for each task using only its own data---each task is associated with a unique online classification model.
	
	\item \textbf{DROM}~\cite{Yang_Li_2020}.
	It is an adaptive primal-dual OMTL algorithm.
	We follow its original setting to set a parameter server but reimplement the communication between workers and the central server asynchronously.
	
	\item \textbf{PMSC-Log}~\cite{Wu_Huang_2016}.
	It is an ADMM-based distributed multi-task algorithm that works under the batch learning setting.
	We modify it to fit the online learning scenario.
	Similar to our approach, PMSC-Log’s objective function combines global and task-specific models.
	However, it does not consider learning the relation among multiple tasks.
	
	\item \textbf{D-PSGD}~\cite{NIPS2017_f7552665}.
	It implements the decentralized parallel stochastic gradient descent in the OMTL setting.
	The step size is set to decrease according to the square of the time step to accelerate the convergence.
\end{itemize}

We follow the original hyperparameter settings in DROM, PMSC-Log, and D-PSGD.
For C-ADMM and D-ADMM, we set $\rho = \lambda_2 = 0.1$, $\lambda_1 = \lambda_3 = \lambda_4 = 0.01$, and $\eta=\sqrt{T}$.
We adopt the cumulative error rate, namely the ratio of the number of mistakes made by an online learner to the number of samples received to date, as a metric for comparing algorithms.
ADMM-Single, DROM and PMSC-Log rely on a centralized parameter server, whereas D-ADMM and D-PSGD are decentralized and will be evaluated using a fully connected and ring topology, respectively.

\subsection{Performance Evaluation.}

Figure~\ref{fig1} depicts the variations of the averaged error rate over the entire online learning process.
Table~\ref{tab:tab2} reports the mean error rates of different algorithms at their last learning round.
The proposed distributed online parallel multi-task learning (C-ADMM and D-ADMM) outperform methods that learn multiple tasks individually with ADMM (ADMM-Single) or learn multiple tasks jointly with optimizers other than ADMM (DROM and D-PSGD) regarding the error rate in most cases.
By comparing D-ADMM with its centralized counterpart, C-ADMM, we observe that implementing ADMM in a decentralized architecture can achieve comparable (or even better) performance than the centralized one, whereas decentralization has scalability benefits in practice.
D-ADMM (Full-Conn) outperforms D-ADMM (Ring) because the fully connected and discretely distributed workers can extract more model information from other peers.
Overall, the proposed C-ADMM and D-ADMM perform better than other baselines.
The C-ADMM excels on the synthetic dataset with the most optimal data distribution, whereas the D-ADMM demonstrates better performance on datasets that more closely resemble real-world scenarios. 
This suggests that D-ADMM is better suited for cases involving unbalanced label distributions and extreme disparities in data quantity among workers.
Furthermore, by comparing the proposed C-ADMM and D-ADMM with PMSC-Log, an ADMM-based OMTL method, without explicitly modeling the task relations, we observe that the former has performance advantages on all datasets.
This result supports our assumption that dynamically modeling task relationships positively affects solving the OMTL problem.

We further evaluate the efficiency of algorithms by setting a target accuracy for each dataset and recording the number of learning rounds and the averaged time consumption per round for each algorithm to reach the accuracy.
Table~\ref{tab:tab3} lists the results.
The plain ADMM-Single has the fastest updating speed but converges to a sub-optimal solution because the ADMM-Single updates the model parameters for each task in parallel on every single worker.
It eliminates the communication overhead caused by parameter sharing among workers. 
However, because of the absence of other task information, each ADMM-Single worker requires more updating rounds to converge and converge sub-optimally, as suggested by Table~\ref{tab:tab2}.
Some OMTL methods (i.e. DROM and PMSC-Log) fail to achieve the specified accuracy.
In comparison, our C-ADMM and D-ADMM have relatively fast convergence and updating speeds.
As stated in~\cite{NIPS2017_f7552665}, the decentralized schema has a comparable computational complexity to the centralized one, but it alleviates the communication overhead of the central servers in the latter.
Considering their advantages in accuracy, as shown in Table~\ref{tab:tab2}, we conclude that the proposed algorithms are efficient and effective for OMTL.

\subsection{Effect of Relationship Learning.}

We further examine the contribution of the proposed relationship learning to the overall OMTL method by conducting an ablation study by removing it and investigating the performance of the remaining parts.
Table~\ref{tab:tab4} lists the variation of the averaged error rate on various datasets of learning tasks independently (denoted as \textit{Indpt}), learning tasks jointly but without relationship modeling (denoted as \textit{W/O RL}) and learning tasks using the proposed methods (denoted as \textit{With RL}). 
The margins in the cumulative error rate demonstrate the effect of the proposed relationship learning module and verify our assumption that modeling relations among tasks is essential for the OMTL problem.
The ablation results on task relationship learning suggest that it will be beneficial for online multi-task learning applications (e.g., in a social search system, searching for creators and for content can be treated as two distinct tasks, and the user experience can be effectively improved through multi-task learning) to learn their implicit relationships.

%% file: sec6.tex
\section{Conclusion}
\label{sec6}

We proposed two distributed OMTL frameworks using a tailored ADMM as the optimizer and an effective mechanism to represent task relations to enhance learning.
The experimental results indicated that the ADMM optimizer, specifically regarding the task relations modeling method, is effective and efficient for learning online-related tasks.
For future work, we wish to extend our methods to multi-class classification settings, which involve evaluating the loss function with multi-class classification mechanisms such as the one-vs-rest strategy.
Furthermore, how to combine the proposed approach with deep learning methods is also worthy of further study. 
In conclusion, our work serves as a beneficial attempt at deriving effective multi-task online learning algorithms for distributed networks.